\documentclass[11pt,a4paper]{article}
\usepackage{times,latexsym}
\usepackage{url}

\usepackage[acceptedWithA]{tacl2021v1}

\usepackage{xspace,mfirstuc,tabulary}

\newif\iftaclinstructions
\taclinstructionsfalse 
\iftaclinstructions
\renewcommand{\confidential}{}
\renewcommand{\anonsubtext}{(No author info supplied here, for consistency with
TACL-submission anonymization requirements)}
\newcommand{\instr}
\fi

\iftaclpubformat 

\else

\fi

\usepackage{colortbl}
\usepackage{times}
\usepackage{latexsym}

\usepackage[T1]{fontenc}
\usepackage[utf8]{inputenc}
\usepackage{microtype}
\usepackage{graphicx}
\usepackage{amsmath}
\usepackage{amssymb}
\usepackage{amsthm}
\usepackage{cleveref}

\usepackage{array}
\usepackage{booktabs}
\usepackage{tcolorbox}
\usepackage{tabularx}
\usepackage{tikz}
\tcbuselibrary{skins}

\usepackage{opensans}
\usepackage[T1]{fontenc}

\usepackage{caption}
\usepackage{subcaption}

\usepackage{newfloat}
\usepackage{caption}

\DeclareFloatingEnvironment[
    fileext=loe,
    listname=List of Examples,
    name=Example,
    placement=tbhp,
]{example}

\DeclareFloatingEnvironment[
    fileext=lot,
    listname=List of Transcripts,
    name=Transcript,
    placement=tbhp,
]{transcript}

\definecolor{headercolor}{RGB}{0,0,0}
\definecolor{bluerow}{HTML}{E6F3FF}
\definecolor{orangerow}{HTML}{FFF4E6}
\definecolor{greyrule}{HTML}{AAAAAA}
\definecolor{linenumcolor}{HTML}{666666}

\newsavebox{\tablebox}
\newcounter{tablelinenum}

\newcommand{\linenum}{\stepcounter{tablelinenum}\thetablelinenum}

\newcommand{\resetlinenums}{\setcounter{tablelinenum}{0}}

\newcommand{\Blue}[1]{%
  \cellcolor{bluerow}\linenum & 
  \cellcolor{bluerow}B: & 
  \cellcolor{bluerow}#1}
\newcommand{\Orange}[1]{%
  \cellcolor{orangerow}\linenum & 
  \cellcolor{orangerow}O: & 
  \cellcolor{orangerow}#1}
\newcommand{\Blueaction}[1]{%
  \cellcolor{bluerow} & 
  \cellcolor{bluerow} & 
  \cellcolor{bluerow}\textit{#1}}
\newcommand{\Orangeaction}[1]{%
  \cellcolor{orangerow} & 
  \cellcolor{orangerow} & 
  \cellcolor{orangerow}\textit{#1}}

\newcommand{\dialoguerule}{%
    \arrayrulecolor{greyrule}%
    \cline{1-3}%
}

\NewEnviron{dialoguetable}[3][]{%
  \def\cornerradius{4pt}
  \def\extraheight{0.6ex}
  \def\linenumwidth{0.2cm}
  \def\speakerwidth{0.2cm}
  
  \ifx&#1&\else
    \foreach \option in {#1} {
      \expandafter\ifx\csname\option @cornerradius\endcsname\relax\else
        \edef\cornerradius{\csname\option @cornerradius\endcsname}
      \fi
      \expandafter\ifx\csname\option @extraheight\endcsname\relax\else
        \edef\extraheight{\csname\option @extraheight\endcsname}
      \fi
      \expandafter\ifx\csname\option @linenumwidth\endcsname\relax\else
        \edef\linenumwidth{\csname\option @linenumwidth\endcsname}
      \fi
      \expandafter\ifx\csname\option @speakerwidth\endcsname\relax\else
        \edef\speakerwidth{\csname\option @speakerwidth\endcsname}
      \fi
    }
  \fi
  
  \addtolength{\extrarowheight}{\extraheight}%
  \arrayrulecolor{greyrule}%
  
  \savebox{\tablebox}{%
    \begin{tabular}{|>{\columncolor{linenumcolor!20}\ttfamily\color{linenumcolor}\centering\small\arraybackslash}b{\linenumwidth}|>{\columncolor{headercolor}\bfseries\raggedright\arraybackslash}b{\speakerwidth} p{\dimexpr\linewidth-\linenumwidth-\speakerwidth-6\tabcolsep-4\arrayrulewidth}|}%
      #2
      \unexpanded\expandafter{\BODY}
      #3
    \end{tabular}}%
  \begin{tikzpicture}
    \begin{scope}
      \clip[rounded corners=\cornerradius] (0,-\dp\tablebox) -- (\wd\tablebox,-\dp\tablebox) -- (\wd\tablebox,\ht\tablebox) -- (0,\ht\tablebox) -- cycle;
      \node at (0,-\dp\tablebox) [anchor=south west,inner sep=0pt]{\usebox{\tablebox}};
    \end{scope}
    \draw[rounded corners=\cornerradius] (0,-\dp\tablebox) -- (\wd\tablebox,-\dp\tablebox) -- (\wd\tablebox,\ht\tablebox) -- (0,\ht\tablebox) -- cycle;
  \end{tikzpicture}
}

\newif\ifanonymous
\anonymoustrue   

\newcommand\blueplayer{{Blue}\xspace}
\newcommand\orangeplayer{{Orange}\xspace}
\newcommand\ak[1]{}
\newcommand\todo[1]{}
\newcommand\jiayi[1]{}
\newcommand\nz[1]{}
\newcommand\enf[1]{}
\newcommand\tea[1]{}
\newcommand\corpus{Portal Dialogue Corpus}

\title{Characterizing Language Use in a Collaborative Situated Game
}

\author{
Nicholas Tomlin$^{*1,2}$,
Naitian Zhou$^{*1}$,
Eve Fleisig$^{1}$,
Liangyuan (Circle) Chen$^{1}$, \\
{\bf
T\'ea Wright$^{1}$,
Lauren Vinh$^{**1}$,
Laura X. Ma$^{**1}$,
Seun Eisape$^{1}$,
Ellie French$^{1}$,
} \\
{\bf
Tingting Du$^{1}$,
Tianjiao Zhang$^{1}$,
Alexander Koller$^{3}$,
Alane Suhr$^{1}$
} \\ 
{
$^{1}$UC Berkeley \hspace{0.2em} $^{2}$NYU \hspace{0.2em} $^{3}$Saarland University
} \\
{
\it
$^{*}$Equal contribution
}
}

\begin{document}
\maketitle
\begin{abstract}
Cooperative video games, where multiple participants must coordinate by communicating and reasoning under uncertainty in complex environments, yield a rich source of language data. We collect the \corpus: a corpus of 11.5 hours of spoken human dialogue in the co-op mode of the popular Portal~2 virtual puzzle game, comprising 24.5K total utterances. We analyze player language and behavior, identifying a number of linguistic phenomena that rarely appear in most existing chitchat or task-oriented dialogue corpora, including complex spatial reference, clarification and repair, and ad-hoc convention formation. To support future analyses of language use in complex, situated, collaborative problem-solving scenarios, we publicly release the corpus, which comprises player videos, audio, transcripts, game state data, and both manual and automatic annotations of language data.
\end{abstract}


\section{Introduction}


Language is the primary medium through which humans coordinate, share information, and achieve common goals. 
To efficiently coordinate in novel contexts, we adapt language to fit our communicative needs under constraints: in jointly embodied environments, we use spatial references whose meanings are dependent on our relative perspectives of the shared scene; in settings with salient novel referents that lack canonical labels, we develop arbitrary but stable and concise conceptual pacts for efficient reference as an interaction proceeds.
Recent work on embodied conversational agents envisions systems that assist or collaborate with human users in situated interactions via language use in context, but such models struggle to adapt their language use to new interaction partners and scenarios.
To build agents capable of efficient, dynamic, and natural interaction in situated environments, we must better understand the linguistic behaviors that support successful coordination between humans. 
This calls for resources that capture, identify, and analyze such behaviors. 

Most existing studies on language-based interaction have been performed on relatively simple game-like environments, where the space of expressible meanings does not change over time and typically reflects a very small subset of real-world meanings; or on open-domain conversational chats, which does not support the fine-grained control of a situated environment and incentive design, nor the capture of the full interaction and its context.
As a result, the set of language phenomena that are used and have been studied in these environments reflects a limited range of linguistic diversity.
Focusing computational studies of language, including the development of language technologies, exclusively on text-based chat or simple situated environments results in neglect of the breadth of possible linguistic devices and the dynamics of their formation and adaptation in interaction.

Multi-agent virtual worlds, like cooperative video games, provide a unique opportunity to study the relationship between the incentives, constraints, and affordances of an interaction's scenario and the interaction's dynamics, including language adaptation and action coordination.
Indeed, game development optimizes player enjoyment via incentive and environment design; games typically come with easily evaluable metrics; and, being virtual, they typically support capturing the entirety of an interaction, rather than only part of it.
Moreover, virtual worlds support a high degree of novelty and pretense for their players~\citep{nguyen2025games}, including fantastical scenarios that we can nonetheless learn to reason and communicate about, such as novel technologies and impossible physics.

\begin{figure*}[h]
\centering
\includegraphics[width=\linewidth]{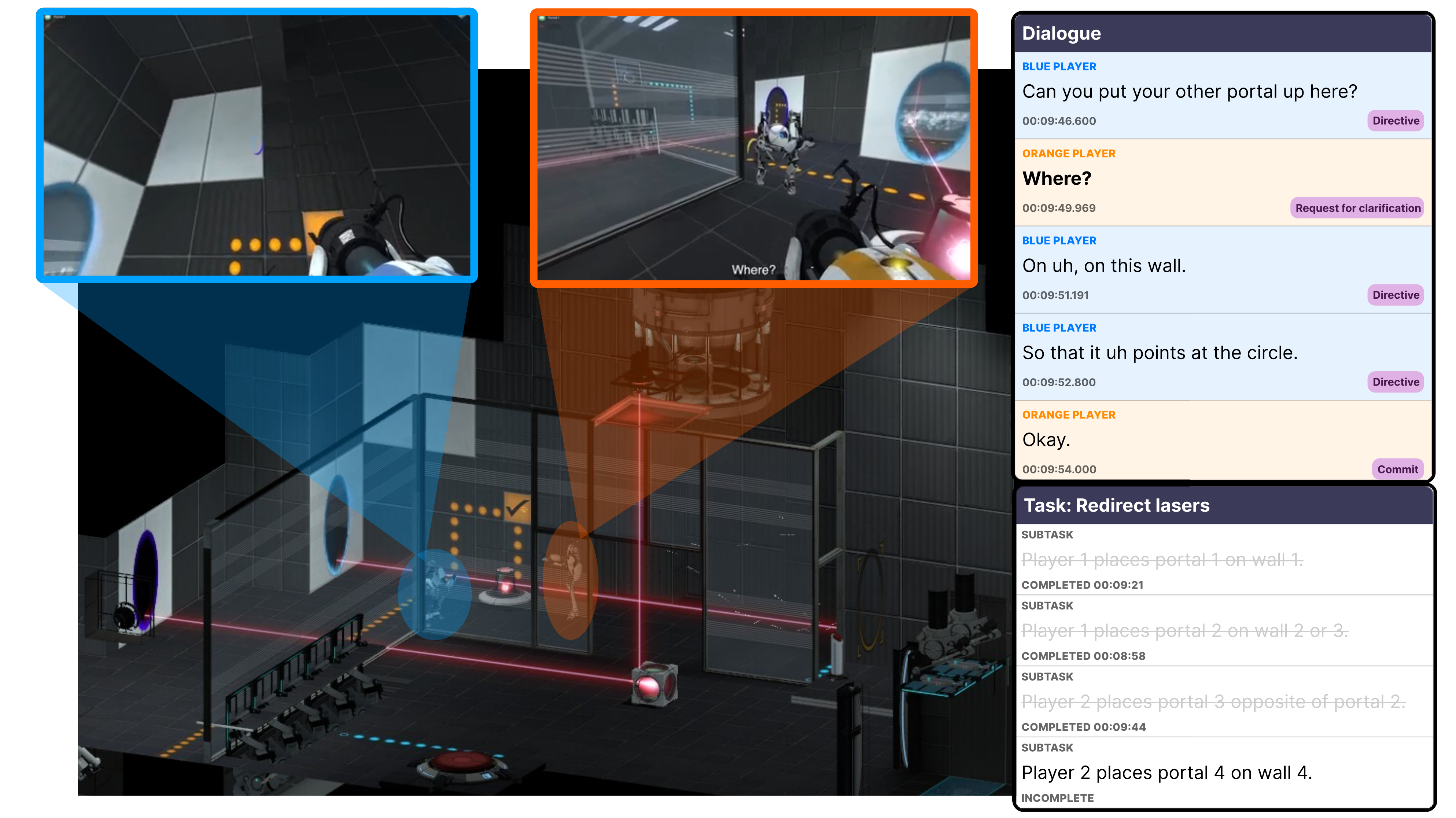}
\caption{Portal~2 is a first-person 3D game where players must solve puzzles through the use of portals. \textbf{(Left)} Snapshot of the game environment, including a third-person view (bottom, not available to players) and popouts showing each player's perspective. \textbf{(Right)} Transcripts of the player dialogue in this scene, each paired with dialogue act annotations (top) and a list of subtasks for this puzzle (bottom, not available to players). In this game snapshot, the blue player uses spatial language to give instructions without an explicit reference, triggering a clarification request and a repair process.  }
\label{fig:screenshot}
\end{figure*}

We study the dynamics of language-based interaction through the introduction of the \corpus: a new corpus of collaborative, situated, and task-oriented human dialogue.
We recorded 11.5 hours of gameplay in the 3D first-person video game Portal~2, where two players collaborate to solve physical puzzles (\Cref{fig:screenshot}).
Players start each puzzle with roughly the same capabilities and prior knowledge, but a high degree of uncertainty about the puzzle solution.
Players must jointly explore each puzzle, including learning new environment dynamics and affordances of novel objects, and construct and execute a solution to the puzzle.
Critically, puzzle designs require that the players coordinate their actions with one another, for example by requiring simultaneous action in different parts of the same puzzle. 
They must therefore communicate, in language, to successfully coordinate under uncertainty.

The \corpus{} combines multiple streams of information, including video and audio recordings of the gameplay, timestamped and transcribed user utterances, and real-time information about the state of the world, including player and object locations and actions. Our dataset's combination of multimodal data, extensive annotation, and inclusion of rich interactions typically absent from conversational datasets makes it a uniquely rich resource for research on task-oriented dialogue, multimodal conversations, and understanding human interaction more broadly.
We demonstrate through quantitative and qualitative analysis of the dataset that players exhibit a number of sophisticated language skills with which current artificial language systems struggle.
For example, players agree on ad-hoc conventions to refer to novel in-game objects or action sequences and use ambiguous spatial references (Section~\ref{sec:reference}). They resolve misunderstandings via generation of clarification requests and self-corrections, and they rely on multimodal grounding, including based on gaze and actions, to resolve ambiguities (Section~\ref{sec:grounding}). Finally, they use task-planning dialogue acts to coordinate  their actions towards successful task completion (Section~\ref{sec:planning}).
The \corpus{} will be made freely available as a contribution to computational research in language and interaction.\footnote{\href{http://berkeley-nlp.github.io/portal-dialogue-corpus/}{\nolinkurl{berkeley-nlp.github.io/portal-dialogue-corpus/}}}\footnote{All videos are available at our YouTube channel \href{https://www.youtube.com/channel/UCQwfmhLbplxf_9b9Hpqb9Tw}{@PortalDialogueCorpus}. Due to our IRB policies, these publicly-available videos do not contain audio, but do contain subtitles. Audio data will be distributed with permission to interested researchers.}

\section{Related Work}



Many existing spoken dialogue datasets comprise open-domain, non-situated conversational speech \cite{switchboard1992, Burnard1995BNC, serban2018survey, doi:10.1126/sciadv.adf3197}. 
In contrast, research on task-oriented dialogue uses settings where agents (either humans or models) must communicate in order to complete shared goals under differences in their knowledge~\citep{allen1995trains, allen2002trips, he-etal-2017-learning, budzianowski-etal-2018-multiwoz, wei-etal-2018-airdialogue, andreas2020dataflow, lin-etal-2024-decision}, or to successfully negotiate under conflicting goals \citep{lewis-etal-2017-deal,he-etal-2018-decoupling,cicero2022}.

In embodied dialogue, agents communicate in the context of partially-observable situated environments. 
Embodiment expands the space of possible asymmetries between agents: in settings with asymmetric action spaces \citep[e.g., CerealBar,][]{suhr2019executing}, collaboration often reduces to instruction-following, and participants need not negotiate their roles via conversation~\citep{narayan-chen-etal-2019-collaborative}; in contrast, tasks with symmetric action spaces where roles are not predefined \citep[e.g., Cards,][]{potts2012goal} are more likely to result in mixed-initiative interactions~\citep{ichikawa-higashinaka-2022-analysis}. 
In our setting, players are not assigned roles ahead of time, which requires them to coordinate on how to divide up their shared goals. 
Like Portal 2, many grounded dialogue tasks are designed to incentivize communication by introducing asymmetry in the agents' observation space \citep[e.g., OneCommon,][]{Udagawa_Aizawa_2019}. 

Other collaborative game settings, such as Hanabi~\citep{bard2020hanabi} or Overcooked~\citep{carroll2019overcooked, NEURIPS2021_797134c3}, have been used to study multi-agent coordination in absence of language-based communication, for example for building agents that model their interaction partners via theory-of-mind.
Each of these domains elicits rich collaborative behavior, but communication is either explicitly limited by game rules or implicitly limited by the space of available meanings. While our setting highlights many similar aspects of collaboration, it focuses more on free-form linguistic communication in an embodied 3D environment, leading to richer reference and conversational grounding phenomena.







\section{The \corpus{}}

Portal is a first-person puzzle video game released by Valve Corporation in 2007.
The key game mechanic is the portal gun: a device that shoots wormhole-like portals onto specific surfaces in the game; once a player has placed two portals, they can teleport between them by entering either portal. 
Puzzle design requires players use portals to reach and move objects to otherwise inaccessible locations.
The game includes several additional mechanics, such as lasers, turrets (defensive robots), and movable bridges.

While Portal is a single-player game, the 2011 sequel Portal~2 includes a two-player cooperative mode called the Cooperative Testing Initiative. This cooperative mode consists of six chapters, each with six or more levels, in which pairs of players must work together to solve puzzles. Each level contains one or two puzzles. In the cooperative mode, shown in \Cref{fig:screenshot}, each player has their own portal gun, resulting in a maximum of four portals that can be placed simultaneously. Most puzzles require both players to place portals, and players can communicate with each other using voice chat. We refer to each player by the color of their respective portal guns: \blueplayer and \orangeplayer.

We collected data of gameplay from the Cooperative Testing Initiative by 18 pairs of players. In total, we obtained 11h 25m of gameplay data from 36 participants, consisting of screen recordings from each player's perspectives, recordings of voice chat audio, and game engine demo files that include exact game state information at each timestep.\footnote{A timestep (tick) is how often the game state is updated, every 1/60th of a second.} The game state includes players' positions, orientations, and objects' locations and velocities. In addition, we produce high-quality, manually-corrected transcripts for the game audio, annotations of dialogue acts, and indicators of when players completed various goals and subgoals within each level.
This allows us to replay each game exactly as it took place during the study, for example to acquire third-person views of gameplay (as in Figure~\ref{fig:screenshot}).
\section{Data Collection}
\label{sec:data}
\subsection{Setup}
We recruited a total of 36 English-speaking participants via flyers, advertisements on online forums, and word-of-mouth.\ifanonymous\footnote{The study was reviewed and approved by our IRB.}
\else\footnote{The study was reviewed and approved by the UC Berkeley CPHS/OPHS under IRB protocol 2023-12-17020.}
\fi
Each pair of players interacted for up to 1 hour each.
We began each session with a written pre-survey, asking participants about their previous experience with video games and the Portal series, as well as consent forms and instructions.
Investigators then guided participants to one of two recording rooms, which were equipped with gaming laptops, as well as identical mice, keyboards, external monitors, and headphones.
Participants played the game in separate rooms and communicated via voice call over Discord. We used Open Broadcaster Software (OBS) to collect audio and screen recordings. 
For each session, we also saved the game engine demo files that contain player and world state during gameplay, from which we can extract in-game events and player positions.
Each pair was instructed to play the first chapter of the game, which consists of six individual levels. Participants who completed the entire first chapter were instructed to signal an investigator, who then reset the recordings and loaded the third chapter, which consists of an additional eight levels.\footnote{We skipped levels in Chapter Two, which we deemed more likely to cause motion sickness due to this chapter's focus on using portals to fall and fly quickly through the air.} No participants progressed to the fourth chapter, and investigators ended each recording session at the end of the hour. Upon completion, each participant was compensated with a gift card.




\subsection{Data Postprocessing}

We used WhisperX~\cite{bain2022whisperx} to generate utterance-level transcriptions from the audio. We then reestimated utterance time alignments using a Wav2Vec-based forced aligner, before manually correcting the utterance segmentation and format in Adobe Premiere. We adapted our transcription format from existing guidelines\footnote{\href{http://ldp-uchicago.github.io/docs/guides/transcription/sect\_4.html}{\nolinkurl{ldp-uchicago.github.io/docs/guides/transcription/sect\_4.html}}} in linguistics to preserve filler words, false starts, and incomplete utterances, as well as extralinguistic communicative expressions like laughter. We delineated utterance boundaries contextually using timing, intonation, and semantic information.
We also anonymized any personal details, such as workplace or hometown, mentioned by participants in the audio recordings and transcripts.\footnote{More detailed information about data post-processing, including manual transcription guidelines and audio and video processing, can be found in the appendix.}

Finally, we hired crowdworkers on Prolific to ensure time alignments of utterances and audio files were high-quality. Workers were paid based on estimated time of completion at a \$15/hour rate. All workers passed an initial quality check and adjustments were spot-checked by project members.

To create videos for each level, we used Adobe Premiere to time-align the screen and audio recordings of each chapter within each pair of players before segmenting recordings into individual levels. 
Audio files were gain-adjusted to normalize the volume across all recordings.

Portal is powered by the Source video game engine, which we configured to generate demo files that include information about the game and player state. We developed a tool that parses demo files to extract, for each timestep (tick) in the game, player position, orientation, and viewpoint information. Demo file timesteps were manually time-aligned to the audio and video recordings.



\subsection{Annotation}
\label{sec:annotation}
After collecting and post-processing interaction recordings,  we generated several layers of metadata relating to dialogue acts and game progress.

\paragraph{Dialogue acts}
To support quantitative analysis of the relationship between language use and player behavior, we release per-utterance tag annotations associated with utterance form, content, and intent.
We designed an annotation schema, adapted from DAMSL (Dialog Act Markup in Several Layers, \citealp{core1997damsl}), that covers five layers of dialogue acts:\footnote{Values and descriptions of the tags are available in Appendix~\ref{app:dialogue_acts}.
}

\begin{enumerate}
\item\textbf{Communicative status} categorizes whether a speaker's intent was successfully communicated or not.
\item\textbf{Information level} describes the type of information conveyed in an utterance. 
\item\textbf{Uncertainty} captures the epistemic stance of the speaker in a given utterance. 
\item\textbf{Utterance type} categorizes spoken dialogue based on its syntactic form.
\item\textbf{Discursive act} describes the function of utterances within a collaborative task-based dialogue. Utterances in Figure~\ref{fig:screenshot} (right) are shown with corresponding discursive acts.
\end{enumerate}
We manually labeled a small subset (three levels; 176 utterances) with four annotators to evaluate both the inter-annotator agreement with our schema.
The annotators first assigned labels given only the text transcript to match the automated classification setting, then performed a second pass with access to the audio and video in order to assess agreement given maximal data. Inter-annotator agreements, calculated as the average pairwise Cohen's $\kappa$, for each layer are presented in Table~\ref{tab:iaa}. We find fair to substantial agreement \cite{landis_measurement_1977} that is roughly in line with previous dialogue act annotations \cite{core1997damsl}.
Low inter-annotator agreement demonstrates the inherent difficulty of the dialogue act annotation task. Somewhat surprisingly, the audio-visual annotations largely did not result in higher agreement compared to the text-only setting; this suggests that, instead of providing disambiguation, access to more modalities might allow for greater flexibility in interpretation. 

To estimate tags for the entire corpus, we used GPT-4o to automatically label each transcribed utterance given its dialogue history, and used the manual annotations to evaluate  the classification performance of GPT-4o (Table~\ref{tab:gpt}). 
To compare agreement levels, we calculate the average agreement between the automated labeling and each of the human annotators.
This agreement is lower than human inter-annotator agreement, suggesting there is still room to improve model performance on this task~\citep{ettinger-etal-2023-expert}.
 


\begin{table}[]
\centering\footnotesize
\begin{tabular}{|l|ccccc|}
\hline
Cohen's $\kappa$ & \textbf{Com.} & \textbf{Inf.} & \textbf{Unc.} & \textbf{Utt.} & \textbf{Dis.} \\
\hline
Text Only & 0.68 & 0.61 & 0.39 & 0.72 & 0.58 \\
\hline
A/V & 0.66 & 0.62 & 0.39 & 0.72 & 0.56 \\
\hline
\end{tabular}
\caption{Pairwise average inter-annotator agreement for when annotators could only access the transcripts and when annotators also had access to the audio and video.}
\label{tab:iaa}
\end{table}

\begin{table}
\centering\footnotesize
\begin{tabular}{|l|ccccc|}
\hline
Cohen's $\kappa$ & \textbf{Com.} & \textbf{Inf.} & \textbf{Unc.} & \textbf{Utt.} & \textbf{Dis.} \\
\hline
GPT-4o & 0.48 & 0.44 & 0.30 & 0.52 & 0.28 \\
\hline
\end{tabular}
\caption{Average agreement between automated labeling and each human annotator.}
\label{tab:gpt}
\end{table}


\paragraph{Tasks and subtasks}

To enable analysis around coordination and joint planning, we identified tasks and subtasks necessary for the completion of each level.\footnote{We limited this annotation to the first chapter because only 12 of the 18 dyads reached chapter three.}
Each puzzle consists of subtasks, defined as any player action necessary for completing the puzzle.
For example, Figure~\ref{fig:screenshot} illustrates a high-level task of redirecting a laser by placing portals in the right configuration; its subtasks include placing all four portals in specific locations.
Some subtasks were completed multiple times within a session, because players would often need to restart part or all of a level in the process of figuring out the puzzle.
Therefore, for each session, we manually recorded the timestamp of the first and last completion of each subtask. 
The generalizability of subtask definitions was validated across multiple sessions with different solutions.
We then grouped subtasks into higher-level tasks: for example, the task of directing a laser to a receptacle might involve two subtasks of placing portals in distinct locations. Task groupings were decided through consensus discussion among authors.











\section{Analysis}

We use our collected corpus and annotations to perform qualitative and quantitative analyses of language use in situated cooperative interaction. 
We first offer descriptive statistics to characterize the task-oriented nature of the data (Section~\ref{sec:data_stats}) before studying specific conversational phenomena like reference (Section~\ref{sec:reference}), conversational grounding (Section~\ref{sec:grounding}), and language use in collaborative planning (Section~\ref{sec:planning}).

\subsection{Data Statistics}
\label{sec:data_stats}

Participants took an average time of 5m~50s to complete each level, although speed varied widely across dyads and levels (standard deviation of 2m~58s). 
12 of the 18 dyads reached the third chapter, with the fastest dyad completing 14 levels. In contrast, the median dyad completed 8 levels, and the slowest completed just 5 levels.
While some pairs were consistently slower than the rest, few dyads were consistently fast at completing levels, and no dyad reached the fourth chapter.



\paragraph{Survey responses} In all sessions, participants did not know each other prior to their participation in the study. Despite this, while most dialogue was task-centered, 843 utterances (3\%) were tagged as non-game-related chit-chat. 
Participants reported varying degrees of familiarity with Portal~2 and video games in general. 
Using a linear regression with the dyad as a random effect, we find that having at least one player with previous exposure to the game is significantly correlated with a better ranking in completion time ($\beta = -6.286; p < 0.05$). 
For each level, we also asked whether players recalled the solution from previous experiences with the game.
Within the dyads that had previously played the game, we find a weak additional correlation between recalling the solution and completion time ranking ($\beta = -3.086; p = 0.067$).


\paragraph{Language statistics} Spoken conversations operate on a precisely-timed, highly-coordinated turn-taking system. Previous works examining the timing of transitions between speakers distinguish between \textit{overlap}, where one speaker begins talking before the previous speaker finishes their utterance, and \textit{gap}, which is a period of silence between the utterance of two different speakers \cite{doi:10.1126/sciadv.adf3197,heldner_pauses_2010}. One result found consistently across languages in spoken conversational dialogue is that the timing of turn transitions is roughly symmetric around 0 \cite{liesenfeld_timing_2023}. Our corpus exhibits greater range of inter-speaker timing than previous conversational corpora, and a larger median gap than overlap. For example, the CANDOR corpus~\cite{doi:10.1126/sciadv.adf3197} includes median gap and overlap lengths of 380ms and -410ms respectively, compared to 666ms and -506ms in our dataset (Figure~\ref{fig:overlap_gaps}). 
This aligns with the fact that, in our setting, participants are performing actions in parallel and in between dialogue acts; gaps may be filled by time where players are attending to the task, environment, and actions, rather than communicating with each other via language. 
Sessions ranged between 577 and 2,045 utterances, with the average being 1,365 utterances per session. Speakers averaged 4.4 words per utterance, resulting in a total of 109K words and 24.5K utterances.\footnote{Additional language statistics, including distributions over utterance lengths, word types, and automatic dialogue act annotations, are available in Appendix \ref{app:lang_stats}.}




\subsection{Reference}
\label{sec:reference}
In an embodied, collaborative setting,
participants must refer to objects, locations, and actions situated within the environment. 
In the \corpus, players not only successfully refer to novel items and actions in the world using a variety of convention formation strategies, but also make pragmatic inferences to resolve different frames of reference under ambiguity.

\paragraph{Convention formation}

In multi-turn interaction, language users form communicative conventions to signal mutual understanding and improve communication efficiency~\cite{krauss1964changes,clark1996referring,fowler1987talkers,garrod1987saying,clark1997conceptual,hawkins2020characterizing,rasenberg2022primacy}.
Prior work has characterized the lifecycle of a convention, from its proposal by one interaction participant, to its refinement and (typically) a reduction in length and complexity through iterated use.
Recent work on modern language models have found that they struggle to participate in this lifecycle in the way human users might expect~\citep{hua2024talk}.
We qualitatively analyze our data for the formation of conventions around both novel and known referents, and for the production of ad-hoc labels of complex action sequences or procedures as a form of shared linguistic abstraction for planning~\cite{mccarthy2021learning,grand2024lilo}.

The Portal game environment introduces players to novel concepts around which they must form conventions.
For example, some game levels include the novel \textit{light bridge} game element, a bridge made of light that extends indefinitely until it hits a solid surface.
Players can use light bridges to cross over hazardous ground surfaces or reach inaccessible locations when portals cannot be placed. 
They can also be used as temporary surfaces that stop the momentum of a moving object, and in some puzzles they must be used in this way to stop a player flying through the air.
In Transcript ~\ref{fig:conventionformation}, \blueplayer uses ``catch'' to refer to the action of creating a light bridge that breaks the fall of another player or object. \orangeplayer adopts the same terminology in a later level, indicating mutual understanding.

\begin{transcript}[h]
    \centering
\begin{dialoguetable}{}{}{}
\Blue{Um, I'll jump on the pad, I'll hit your -- uh, I'll hit the bridge, and then you place the (bridge).} \\
\dialoguerule
\Orange{Yeah.} \\
\dialoguerule
\Blue{And it, like, sort of \textbf{catches} me.} \\
\dialoguerule
\Orange{Yeah.} \\
\dialoguerule
\dialoguerule
\Orangeaction{...In a later level:} \\
\Orange{You might be able to \textbf{catch} it.} \\
\end{dialoguetable}
    \caption{An example of convention formation.}
    \label{fig:conventionformation}
\end{transcript}

We also find instances of metonymy in our dataset; metonymy is a phenomenon in which a word or phrase is used to refer to something related, but not identical, to its literal meaning ~\cite{lakoff2024metaphors, kovecesandradden1998metonymy, alavc2004man, littlemore2015metonymy}. In Transcript ~\ref{fig:metonymy}, \blueplayer refers to portals by their color alone, shortening the reference based on salient features.

\begin{transcript}[h]
    \centering
\begin{dialoguetable}{}{}{}
\resetlinenums
\Blue{Y-- you can’t, can't you shoot your, your \textbf{orange} on top of my \textbf{blue}?} \\
\end{dialoguetable}
    \caption{An example of metonymy.}
    \label{fig:metonymy}
\end{transcript}

Beyond creating shared labels for objects and actions, we also observe dyads creating abstractions for more complex or procedural concepts. On-the-fly abstraction occurs when a player spontaneously references a strategy, concept, or sequence of actions with a concise phrase. In Transcript ~{\ref{fig:ontheflyabstraction}, ``the same trick'' is used by \blueplayer when suggesting they reattempt to execute a plan.
Of course, this abstraction cannot refer to the exact sequence of actions previously taken by the players, because reattempting those would lead to the same failure as before. 
Instead, this abstraction relies on the players' mutual knowledge of a joint plan and its variations, including potential mistakes to avoid.


\begin{transcript}[h]
    \centering
\begin{dialoguetable}{}{}{}
\resetlinenums
\Orangeaction{\orangeplayer mistakenly enters portal.} \\
\Orange{What?} \\
\Orange{Oh.} \\
\hline
\Blue{Alright.} \\
\Blue{It’s -- I mean, it's okay, so --} \\
\hline
\Orange{I gotta go through.} \\
\hline
\Blue{Uh, yeah, let's do -- do -- do \textbf{the same trick}.} \\
\end{dialoguetable}
    \caption{An example of on-the-fly abstraction.}
    \label{fig:ontheflyabstraction}
\end{transcript}

\paragraph{Spatial reference}
\begin{figure*}[h]
    \centering
    \includegraphics[width=1.\linewidth,page=1,trim=0 745 0 0,clip]{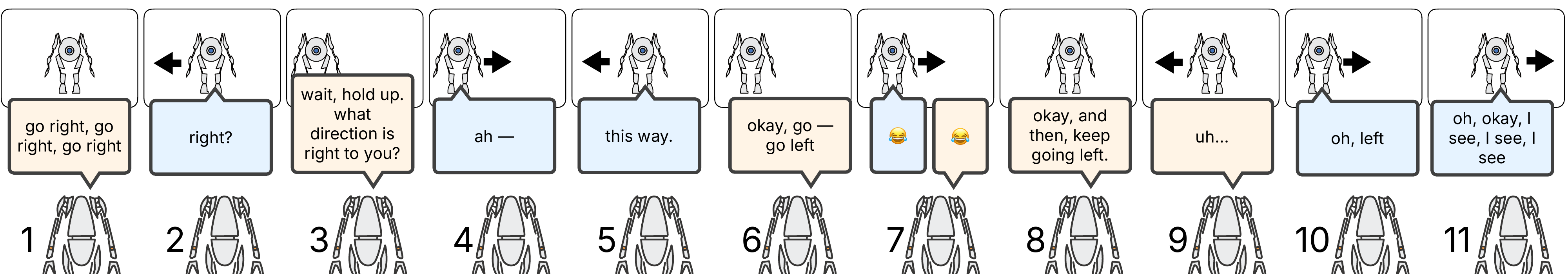}
    \caption{Here, \blueplayer (top) and \orangeplayer (bottom) are facing one another, and \blueplayer is stuck in a box out of which  \orangeplayer is guiding them.}
    \label{fig:spatial_portal}
\end{figure*}
When participants of an interaction are jointly embodied in the same environment but occupy different bodies, they necessarily have different perspectives and observations. 
To successfully make and resolve references, each participant must resolve uncertainty over the others' observations of the shared world, and whose perspective they are taking during communication~\citep{Levinson2003SpaceIL}.
In 3D environments like Portal 2, this gives rise to spatial language with ambiguous frames of reference, which existing vision-language models struggle with~\citep{tang-etal-2024-grounding}.
Figure~\ref{fig:spatial_portal} shows two players jointly resolving uncertainty about the meaning of individual spatial referring expressions, as well as the frame of reference being used to produce them.
Here, \orangeplayer and \blueplayer face one another, and \orangeplayer attempts to guide \blueplayer out of a box in which \blueplayer is stuck.

\orangeplayer begins by instructing \blueplayer to \textit{go right} (Turn 1), which immediately introduces ambiguity over the underlying reference frame: right of what?
\blueplayer attempts to disambiguate by asking for confirmation that this means they should move to their literal right; i.e., that \orangeplayer was using a listener-centered perspective (Turn 2)~\citep{SCHOBER19931,DINGEMANSE202430}. 
\orangeplayer then explicates this ambiguity by asking \blueplayer to clarify their frame of reference (Turn 3).
To repair this uncertainty, \blueplayer demonstrates \textit{right} to \orangeplayer, by moving right while labeling the action with \textit{this way} (Turn 5)~\citep{Keevallik01012013}.
This prompts \orangeplayer to match their frame of reference to \blueplayer's assumption (Turn 6), and both players acknowledge the successful repair through laughter (Turn 7)~\citep{shaw2013,koivisto2019}.
\orangeplayer continues by instructing \blueplayer to \textit{keep going left} (Turn 8), but \blueplayer has over-corrected and adopted \orangeplayer's initial egocentric frame of reference, moving to \orangeplayer's literal left instead (Turn 9).
\orangeplayer draws \blueplayer's attention to this mistake via a hesitation (\textit{uh...}), and \blueplayer responds by confirming that spatial references made by \orangeplayer should be interpreted literally (Turn 10).
Finally, \blueplayer's last utterance can be interpreted as an acknowledgment that the pair has established a convention on the spatial frame of reference (Turn 11)~\citep{SCHOBER19931}.

\subsection{Conversational Grounding}
\label{sec:grounding}

Successful interaction between interlocutors requires the establishment of a shared understanding, or common ground, through the process of conversational grounding \cite{clarkGroundingCommunication1991}.
Grounding often occurs through multiple turns as ambiguities are negotiated \cite{benottiGroundingCollaborativeProcess2021}, and may require that interlocutors draw on other resources or modalities beyond language \cite{goodwin_action_2000,mohapatraConversationalGroundingAnnotation2024}.
Recent work has found that existing language models struggle to drive forth conversation via grounding~\citep{shaikh-etal-2024-grounding}.
The \corpus{} includes rich examples of multi-turn, multi-modal conversational grounding.

\paragraph{Clarification requests}

\resetlinenums
\begin{transcript}[h]
    \centering
\begin{dialoguetable}{}{}{}
\Blue{Can you put your other portal up here?} \\
\Blueaction{\blueplayer faces wall.} \\
\hline
\Orange{Where?} \\
\hline
\Blue{On uh, on this wall.} \\
\hline
\Blue{So that it uh points at the circle.} \\
\hline
\Orange{Okay.} \\
\end{dialoguetable}
    \caption{A simple clarification request.}
    \label{fig:clarification}
\end{transcript}

Even a simple clarification request may require the construction of a \textit{subdialogue} with multiple turns.  
In Transcript~\ref{fig:clarification} (reproduced from \Cref{fig:screenshot}), \blueplayer issues a directive to \orangeplayer (Line 1).
However, \orangeplayer does not take up this utterance as a directive, which would elicit an action, an acceptance, or a rejection; instead, \orangeplayer asks for clarification (Line 2). Only after \blueplayer resolves this ambiguity does \orangeplayer accept the original directive (Line 5).


\begin{transcript}[h]
    \centering
\begin{dialoguetable}{}{}{}
\resetlinenums
\Blue{I think that's what we're supposed to do.} \\
\hline
\Orange{Huh?} \\
\hline
\Blue{At the end of each level, we're supposed to explode.} \\
\end{dialoguetable}
    \caption{An open-ended clarification request.}
    \label{fig:clar_uh}
\end{transcript}

Our automated annotations label 532 utterances as requests for clarification. 
While Transcript~\ref{fig:clarification} features a class-specific clarification in which \blueplayer is constrained to responding with a location \cite{dingemanse_formats_2014}, our dataset also includes more ambiguous, open-class clarification requests \cite{drew_open_1997,raymond_huh_2013}. For example, in Transcript~\ref{fig:clar_uh}, \orangeplayer says ``Huh?'' (Line 2), which is sufficient for \blueplayer to clarify their statement.

\paragraph{Repair}

While recent work in natural language processing has focused on building systems that establish conversational ground by making clarification requests \cite{testoni_asking_2024,hou_decomposing_2024,zhang_clarify_2025}, we can view clarification requests as a special case of repair more broadly. In conversation analysis  literature, requests for clarification are often referred to as ``other-initiated self-repair'', in which the repair is prompted for by the \textit{other} interlocutor \cite{kendrick_other-initiated_2015,bostrom_other-initiated_2021}.

\begin{transcript}[h]
    \centering
\begin{dialoguetable}{}{}{}
\resetlinenums
\Orange{And then shoot my yellow (one) -- actually, my red one with your other one.
} \\
\end{dialoguetable}
    \caption{Example of self-initiated self-repair.}
    \label{fig:selfrepair}
\end{transcript}

Other-initiated repair is common, and often the only available kind of repair, in most chatbot contexts because text-based chat interfaces  enforce sequential turn-taking. However, self-repair has been found to be preferred in naturalistic English conversation \cite{schegloff_preference_1977}, and our dataset also includes many instances of self-initiated repair (cf. Transcript~\ref{fig:selfrepair}), including 1,922 utterances automatically tagged as correction.

\paragraph{Multimodal conversational grounding}
In Portal 2, like other embodied environments, interaction participants attend to each other as well as their environment and tasks \cite{holt_interactive_2006,goodwin_action_2000}. 
We find that in addition to spoken communication, players use extralinguistic resources such as gaze, movement, and object-use to assist in establishing a common ground.
For example, in Transcript~\ref{fig:clarification} (from Figure~\ref{fig:screenshot}), \blueplayer faces a wall on which a portal can be placed, and asks \orangeplayer to place a portal ``here'', expecting \orangeplayer to resolve the referent of ``here'' using their observation of \blueplayer's gaze.
While this first attempt at reference fails, the ensuing dialogue, combined with embodied action including \blueplayer jumping while facing the wall, allows \orangeplayer to successfully resolve its meaning.


\begin{transcript}[h]
    \centering
\begin{dialoguetable}{}{}{}
\resetlinenums
\Orangeaction{\orangeplayer shoots portal at lever.} \\
\Orange{Right here, uh, behind you.} \\
\hline
\Blueaction{\blueplayer turns around.} \\
\Blue{Oh.} \\
\end{dialoguetable}
    \caption{Shooting portals sometimes constitutes a pointing gesture.}
    \label{fig:deixis}
\end{transcript}

In embodied interaction, gestures often co-occur with deictic expressions \cite{mondada_challenges_2016}.
Portal 2 limits player movements to turning, moving forward and backward, crouching, jumping, and shooting the portal gun; they cannot move limbs independently, for example to point.
We find that, in absence of the ability to literally point, many players shoot the portal gun at in-game objects or locations to serve as a pointing gesture, relying on the visual animation that plays when the portal gun is shot on a surface. Transcript~\ref{fig:deixis} shows how \orangeplayer combines deixis (``Right here'') with gesture (shooting the portal gun at the lever). 

Finally, while speech is usually sequentially organized through precise turn-taking mechanisms, simultaneous choral speech (where multiple speakers speak at the same time) can be deployed for a variety of ends \cite{pfander_turn-sharing_2019,mondada_sequential_2025}. This is especially relevant in multimodal settings due to the co-temporality of language and action \cite{mondada_multiple_2018}. In our data, players demonstrate task-oriented simultaneous speech by counting aloud to synchronize actions like movement, portal placement, and pressing buttons or pulling switches.

\subsection{Collaborative Problem Solving}
\label{sec:planning}

Research in collaborative problem solving studies how people communicate, make plans, and resolve uncertainty in order to complete complex tasks.
Prior work has studied language-based collaboration in cooperative tasks like problem-solving and learning~\citep{Grosz1977representation,GROSZ1996269,puntambekar2006analyzing,baker2015collaboration, graesser2018advancing,bara-etal-2021-mindcraft,jeknic-etal-2024-dialogue,jeknic-etal-2025-collaborative}.
Our setting emphasizes the embodied and dynamic nature of the interaction, where players iteratively and jointly construct possible puzzle solutions, execute actions that test out these solutions, and reformulate their joint understanding of the puzzle.



\paragraph{Mixed-initiative interaction}
In our setting, players are not assigned roles a priori.
This means there are no specific incentives for imbalanced communication, for example where one player simply gives instructions to the other~\citep[like in CerealBar, ][]{suhr2019executing}.
Despite this, dyads varied widely in how they distributed communication among themselves (Figure~\ref{fig:cum_utterances}).
For example, in Session~1, \blueplayer spoke for 20m~54s while \orangeplayer only spoke for 3m~33s; additionally, based on our automatic analysis, 19.8\% of \blueplayer's utterances were directives, while only 2.5\% of \orangeplayer's utterances were directives.
Most dyads exhibited more balanced communication, with the median difference in speaking time between players at only 4m 8s.
Within  more balanced dyads, we observe several common communicative strategies. 
Qualitatively, we find that players stuck on a puzzle often took turns describing their observations or affordances of the environment. 
Additionally, one or both players often narrated their own actions, as described in \citet{roschelle1995construction}, even if a joint plan wasn't yet clear. 
Once players converged upon a solution, it was relatively rare for them to communicate explicitly about their plan, instead relying on the shared understanding built up over the course of the interaction.


\paragraph{Progress tracking and task difficulty}

\begin{figure*}[t]
\includegraphics[width=\linewidth]{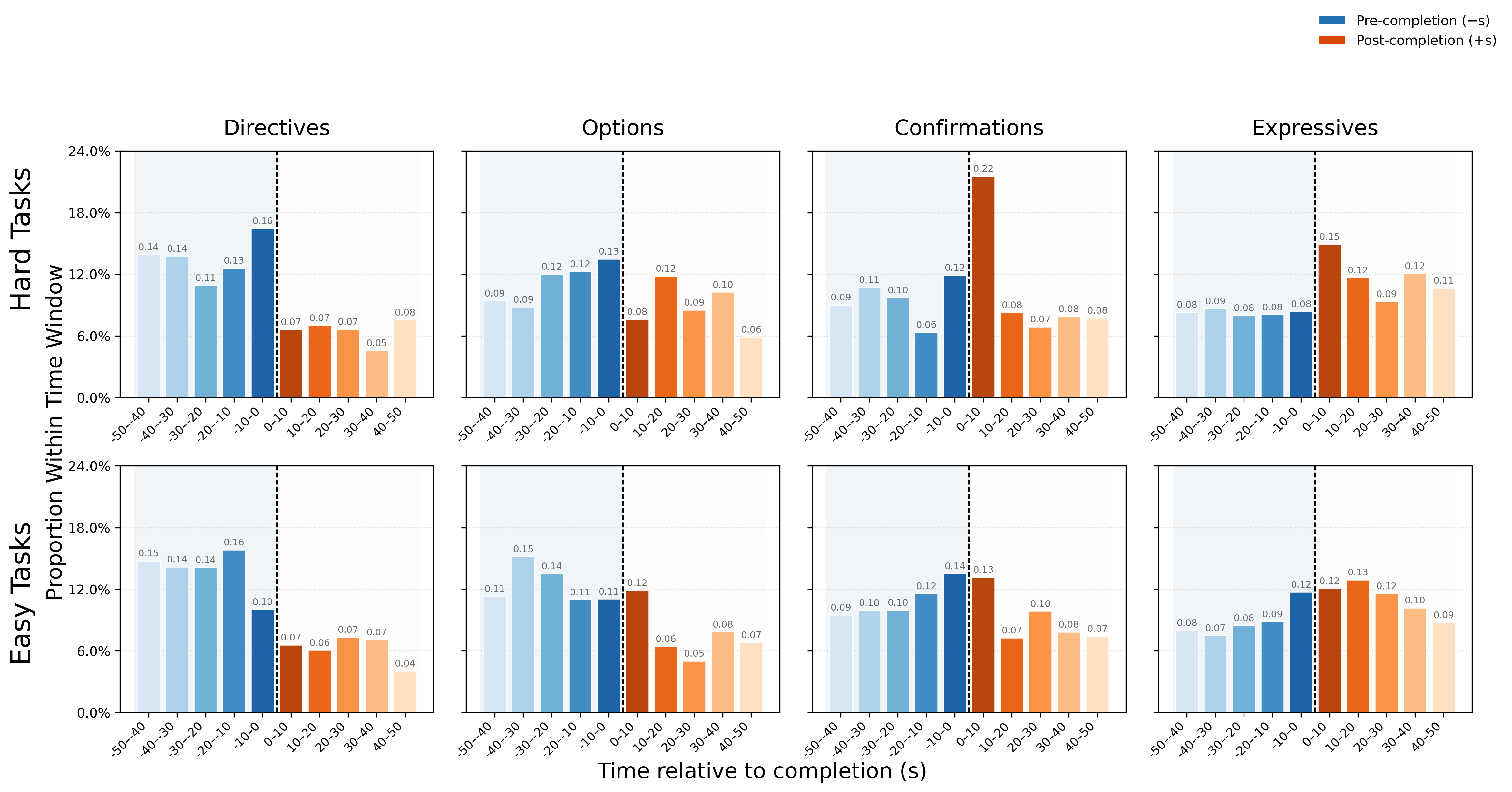}
\caption{Proportion of each discursive act within the 50s before or after task completion. The top row is calculated on the five most difficult tasks, and the bottom row on the five easiest for comparison.}
\label{fig:subtasks}
\end{figure*}

In dynamic and situated environments, collaborative problem solving often requires monitoring the progress of the task at hand. 
To investigate possible linguistic markers of task completion, we compare dialogue patterns in easy versus hard tasks. 
Using annotated task timestamps (Section~\ref{sec:annotation}), we compute the task completion time for all coarse-grained tasks across the first six levels by subtracting the timestamp from the first attempt of the first subtask from the final attempt of the final subtask.
Across all dyads, tasks were completed in 53s on average, with the hardest five tasks averaging 2m~14s and the easiest averaging only 3s.
We then consider the five tasks with the longest average completion time as the most difficult tasks, and the five tasks with the shortest completion time as the easiest tasks.
Difficult tasks, like jointly navigating a maze (average time: 3m 3s) and placing four portals to precisely align a laser (average time: 2m 1s), require sophisticated coordination.
Easy tasks, like placing an object in a specified location (average time: <1s) or placing two portals to allow oneself to move to an inaccessible location (average time: 0.21s), can be done individually without communication.

We hypothesize that the language use surrounding these task executions varies significantly depending on task difficulty.
Because task times vary significantly across dyads, especially for difficult tasks, we focus on analyzing language used around the completion of each task.
We consider the utterances 50 seconds before and after completion of each task for all dyads, and divide these into ten sets of utterances (five before task completion, five after).
In Figure~\ref{fig:subtasks}, we compare the prevalence of four discursive acts (Section~\ref{sec:annotation}) in the pre-completion utterance sets to its prevalence in the post-completion sets. 
The dynamics of discursive acts changes significantly based on task difficulty, especially within the ten seconds before and after task completion.
In difficult tasks, confirmations (e.g., ``Cube acquired.'') increase by 83\% immediately after completion. 
Directives (e.g., ``Okay, step on the box'') drop by 56\% and offers and options (e.g., ``And then I can lower you down.'') by 38\%, demonstrating a shift away from planning and problem-solving language once the goal is achieved. 
Finally, expressives (often celebrations of success, including laughing) increase by 88\%. 
Across discursive acts, easy tasks show less stark or immediate changes with decreases all less than 30\%.
However, in both easy and hard tasks, analysis of the information-level layer tags shows that non-task-related speech increases significantly after task completion, with a 200\% increase in difficult tasks and a 400\% increase in easy tasks (not plotted). 
Shifts in language use allow us to identify when dyads transition from problem solving to completion, especially on difficult tasks. The relative weakness of these indicators in easy tasks implies that problem-solving language and progress monitoring emerge most strongly when tasks require sustained collaborative effort.

\section{Conclusion}
The \corpus{} comprises extensively annotated human dialogue in a situated, collaborative interaction scenario.
This corpus allows us to study linguistic phenomena in a complex, goal-oriented setting.
We find that players engage in varied and understudied linguistic strategies: convention formation, spatial reference, conversational grounding from multimodal signals, and frequent collaborative planning. This wide range of linguistic behaviors that emerges in complex interactions helps to illuminate crucial challenges of language use in collaborative problem solving: both for understanding language use in these interactions among people, and for the development of AI systems used in complex multimodal settings. We provide the dataset as a resource for future work on understanding language in interaction. 
\newpage

\section*{Acknowledgments}
We would like to thank Elisabeth Andr\'{e}, Justine Cassell, William Chen, Jonathan Ginzburg, Robert Hawkins, Syrielle Montariol, Jiayi Pan, Massimo Poesio, Matt Purver, Edwin Sipson, Zineng Tang, David Traum, and Kayo Yin for their insights, feedback, and discussion.
NZ, EF, and TW were supported by funding from the National Science Foundation (Graduate Research Fellowship DGE-2146752).
This work was also supported by an Ai2 Young Investigator award.

\bibliography{anthology,custom}
\bibliographystyle{acl_natbib}
\clearpage
\appendix

\section{Additional Language Statistics}
\label{app:lang_stats}
Figures~\ref{fig:zipf}, \ref{fig:utterance_lengths}, \ref{fig:overlap_gaps}, \ref{fig:cum_utterances} and \ref{fig:dialogue_distributions} give additional language statistics. Token length, gap length, and overlap length all follow exponential distributions, with fewer tokens, smaller overlaps, and smaller gaps being most common.

\begin{figure}[h]
    \centering
    \includegraphics[width=0.85\linewidth]{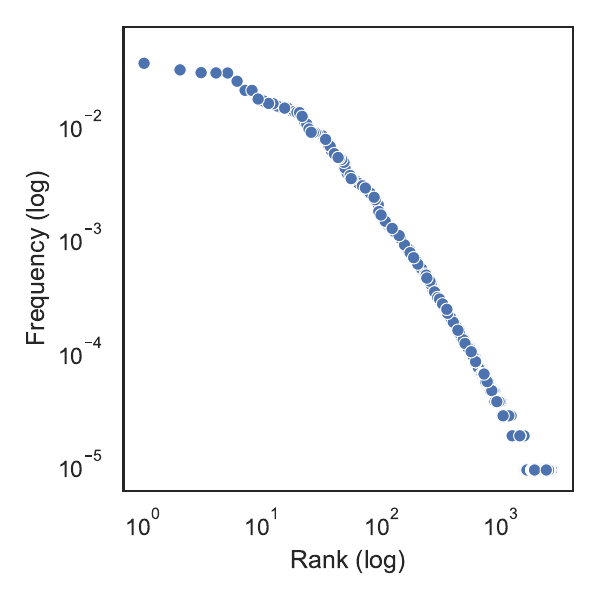}
    \caption{The token frequencies in our corpus follow a Zipfian distribution.}
    \label{fig:zipf}
\end{figure}

\begin{figure}[h]
    \centering
    \includegraphics[width=0.9\linewidth]{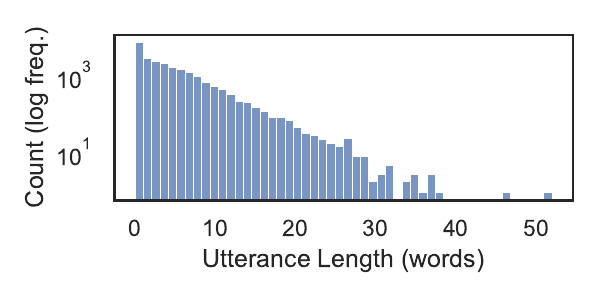}
    \caption{The token length (number of words per utterance) follows an exponential distribution.}
    \label{fig:utterance_lengths}
\end{figure}

\begin{figure}
    \centering
    \includegraphics[width=0.9\linewidth]{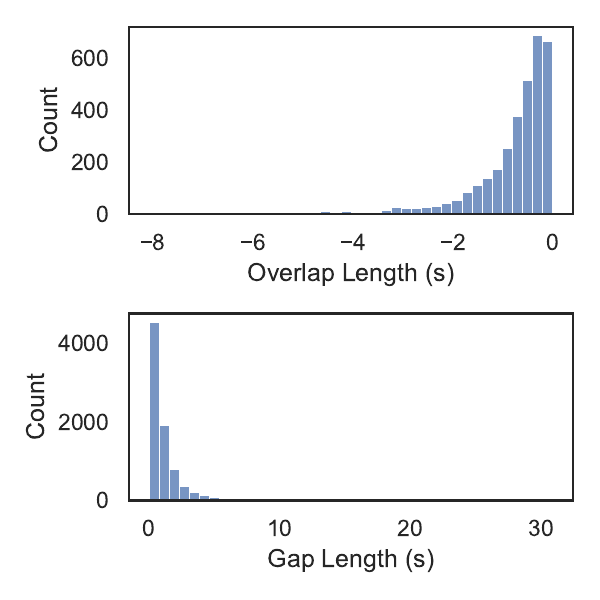}
    \caption{Distribution of gap and overlap lengths: both follow exponential distributions that peak close to length 0. Drawing from \citet{heldner_pauses_2010}, we define the gap as the duration of silence between the end of one speaker and the beginning of the next speaker. Similarly, we define the overlap as the duration of speech between the beginning of the next speaker and the end of the previous speaker, if one begins before the other finishes. We exclude cases when the speaker of an utterance is the same as the speaker of the next one.}
    \label{fig:overlap_gaps}
\end{figure}

\begin{figure}
    \centering
    \includegraphics[width=1\linewidth]{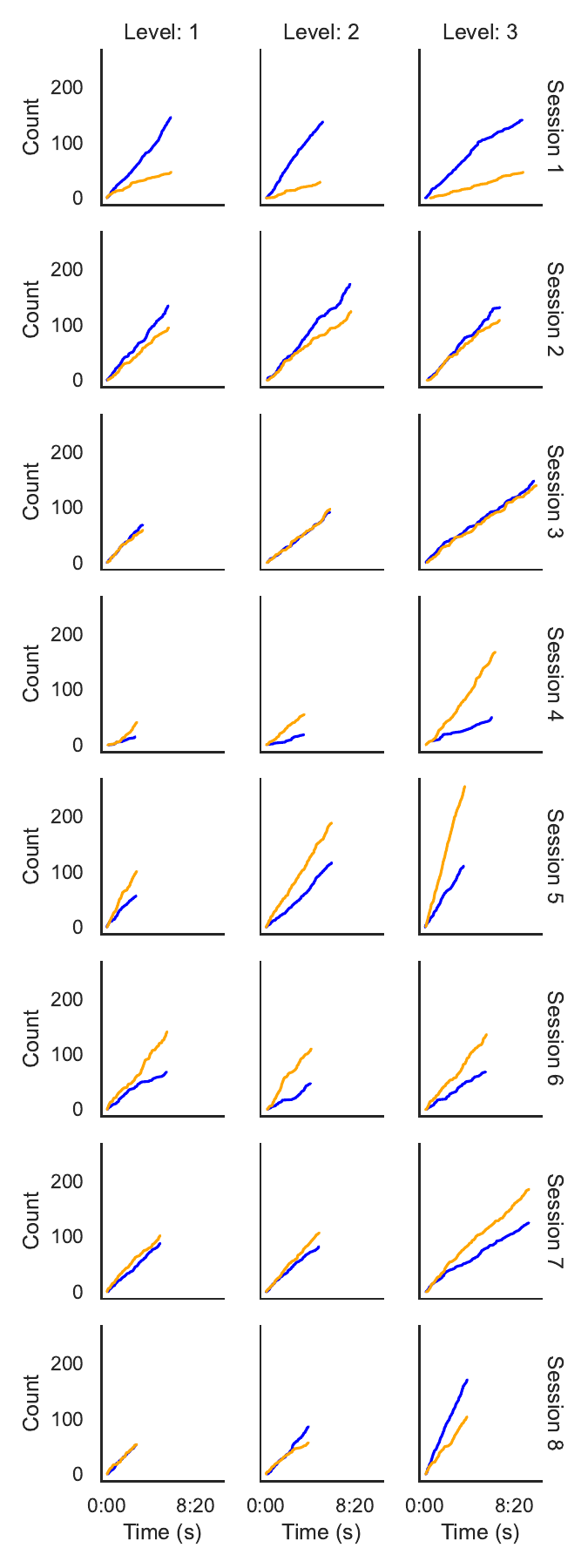}
    \caption{The cumulative number of utterances by each player over the course of the first three levels in the first 8 sessions. Dyads vary in the extent to which their communications are balanced; e.g., relatively unbalanced in session 1, and relatively balanced in sessions 2 and 3.}
    \label{fig:cum_utterances}
\end{figure}

\begin{figure}
    \centering
    \begin{subfigure}[b]{0.32\textwidth}
        \centering
        \includegraphics[width=\textwidth]{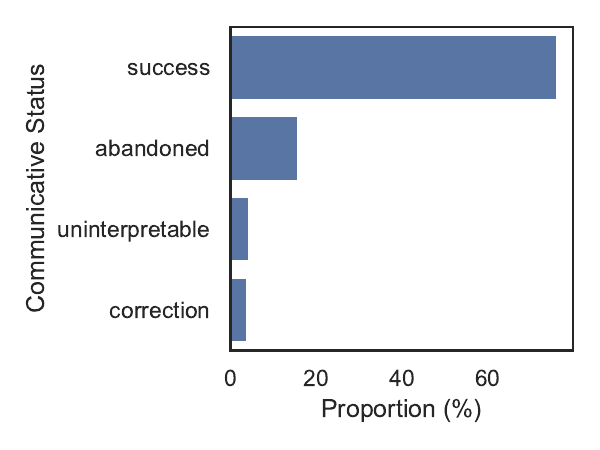}
        \vspace{-2em} \caption{Communicative Status}
        \label{fig:comm_status}
    \end{subfigure}
    \hfill
    \begin{subfigure}[b]{0.32\textwidth}
        \centering
        \includegraphics[width=\textwidth]{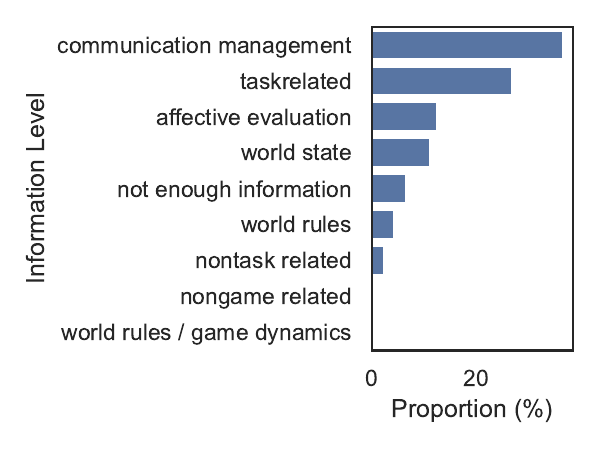}
        \vspace{-2em} 
        \caption{Information Level}
        \label{fig:info_level}
    \end{subfigure}
    \hfill
    \begin{subfigure}[b]{0.32\textwidth}
        \centering
        \includegraphics[width=\textwidth]{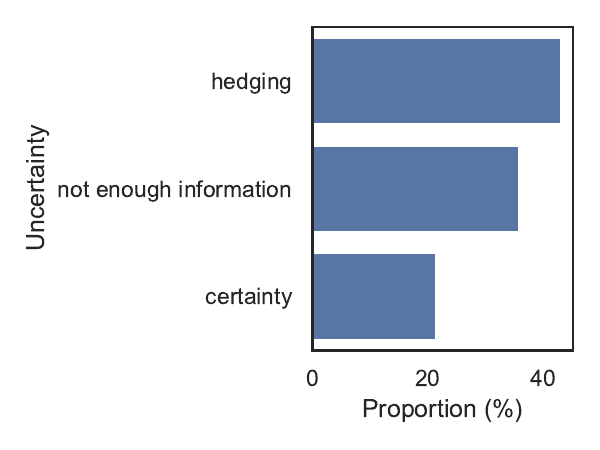}
        \vspace{-2em} 
        \caption{Uncertainty}
        \label{fig:uncertainty}
    \end{subfigure}
    
    \begin{subfigure}[b]{0.31\textwidth}
        \centering
        \includegraphics[width=\textwidth]{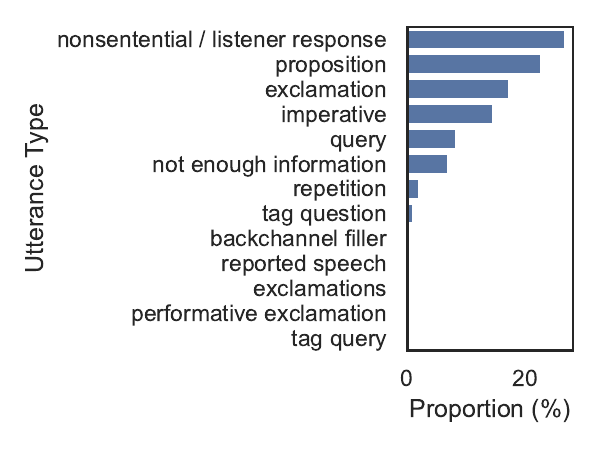}
        \vspace{-2em} 
        \caption{Utterance Type}
        \label{fig:utterance_type}
    \end{subfigure}
    \begin{subfigure}[b]{0.32\textwidth}
        \centering
        \includegraphics[width=\textwidth]{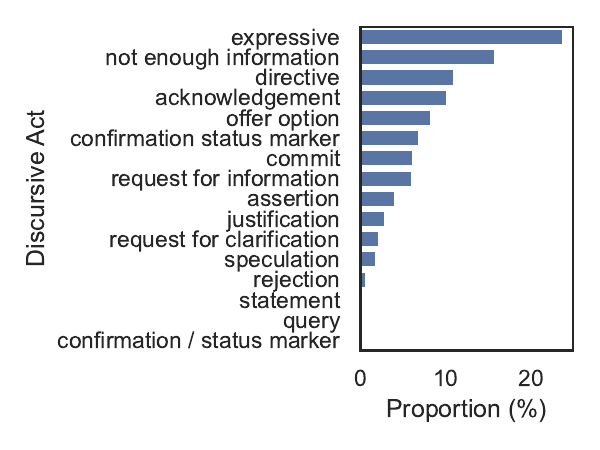}
        \vspace{-2em} 
        \caption{Discursive Act}
        \label{fig:discursive_act}
    \end{subfigure}
    \hfill
    \caption{Distribution of dialogue annotations in the data. Most interactions are successful, though >15\% are abandoned. Most utterances relate to managing communication and task-related issues. >40\% of utterances include hedging. Listener responses and expressive acts are the most common utterance types and discursive acts, respectively.}
    \label{fig:dialogue_distributions}
\end{figure}

\section{Pre-Gameplay Survey and Instructions}
\label{app:instructions}

\subsection{Preparing screen and audio recordings}

We used OBS to record gameplay footage and capture microphone input. We recorded at 30 frames per second with 1920$\times$1200 resolution. The audio was recorded from the headset microphone in mono channel with a 48K sampling rate.

\subsection{Configuring Portal}

We make several configuration changes to the Portal game. We disable the visual pointing and partner view (where the player can see the other players' view) features to encourage planning and reference through language use. We also map the ``L'' key to trigger a ``mark'' event that is later used to align the game state recordings with the audio and screen recordings.

We disable in-game sounds and captions for a cleaner recording.

\subsection{Day-of protocol}
Two facilitators were present at each session. They were provided with a script to follow to introduce the session to the participants.

\newcommand{\scriptmeta}[1]{\textit{\textcolor{gray}{#1}}}

\subsubsection{Script}
\paragraph{Introducing participants} Welcome to the collaborative video game study!  Let’s introduce ourselves. My name is \scriptmeta{facilitator 1 name} and this is \scriptmeta{facilitator 2 name}.

\scriptmeta{[Have participants introduce themselves to one another]}

First you will fill out a couple of forms, and then we’ll get started.

\paragraph{Signing forms and survey}

\scriptmeta{[Hand out consent forms]}

We have two consent forms for you to fill out. Your consent to both is required to participate in the study.
The first form is for your informed consent to participate in our study. It describes the study procedures, benefits and risks, confidentiality, and a few other details. Please read through this and sign if you consent to participate. 

The second form is a media records release form. It describes the data we will collect during the study, and how we plan to use it. Please read through it and sign if you consent to the records release. 

If you have any questions about these forms, feel free to ask.

\scriptmeta{[Hand out survey]}

We also have this survey we would like you to fill out that asks about your experiences with video games and language. Please fill it out and let us know when you’ve completed it. Let us know if you have any questions about it.

\paragraph{Protocol description}
For the next hour or so, until \scriptmeta{time when session is up}, you’ll play the Portal 2 co-op game starting with the introductory levels and moving onto more challenging levels.

\scriptmeta{[Hand out levels tracking sheet]}

This sheet will guide you through the levels of the game. Right now, we’ll start with the first course, which is an introductory set of six levels. We are skipping the 2nd course of the game.

You are not expected to get to all of the game levels! The whole game takes about five hours in total. We are just listing all of the valid levels for completeness.

At the beginning of each level, one or both of you should press the ``L'' key. Make sure both of you see the word ``MARK'' pop up on the screen. This is to let us align various recordings together.

After each level, please check off whether this level was new to you (which would be true for people who have never played before), or if you remembered the solution from playing the game previously, please check that off too.

After every set of levels, you will take a break. This sheet has reminders about when courses are complete. We will be sitting outside of the rooms working on other things, so when this happens, please exit your rooms and let us know so we can stop recordings and you can take a short break.

In the first course, if you and your partner become completely stuck because the mechanics are confusing, please let us know by finding us outside of the rooms, and we can offer assistance.

\scriptmeta{[Hand out mechanics sheet]}

Here’s a short tutorial sheet that describes the game controls. If you are unfamiliar with the game, please take a brief look at it and let us know if you have any questions.

If you have played the game before, we’ve disabled two of the game features you might be familiar with: the ping tool and the ability to see the other person’s perspective.

We also turned off the sound and captions in the game.

\paragraph{Moving to each room}

Ok, let’s get started! \scriptmeta{Participant 1 name} let’s go to the other room.
I’ll start the recording.
\begin{itemize}
    \item  \scriptmeta{Press Ctrl+Shift+F11 to start recording video/audio}
    \item \scriptmeta{Press the windows key and check that there is a red circle on the OBS icon to confirm that it is recording.}
    \item \scriptmeta{Press \texttt{\textasciitilde{}}, run \texttt{record~<demofilename>}. The \texttt{demofilename} should be \texttt{sessionID\_uniqueID\_course1}}
    \item \scriptmeta{Press L to mark time}
    \item \scriptmeta{Walk to and enter level 1}
\end{itemize}
Here is your seat. Here’s the keyboard and mouse, monitor, and headset. First, can you test whether you can hear each other? Then you should be good to start playing!

\paragraph{After each course}

\begin{itemize}
\item \scriptmeta{Stop the A/V recording - OBS is configured with CTRL-SHIFT-F12 to stop recording}
\item \scriptmeta{Stop the demo - open dev console with \texttt{\textasciitilde{}}, run \texttt{stop}}
\item \scriptmeta{Check whether data is successfully saved and stored to SSD}
\end{itemize}

\section{Transcription Guidelines}
\label{app:transcription}

The transcripts of players' speech were manually corrected and standardized according to these guidelines.

\subsection{Utterance Segmentation}

To determine whether two segments should be two utterances or merged as one:
\begin{enumerate} 
\item Follow the 2s rule
\item Use changes in intonation and what we can infer about the semantics to decide whether to split for shorter pauses
\begin{enumerate}
\item Descending pitch can be a hint that it is an utterance ending
\item If it would be difficult to assign different functional meanings to two adjacent utterances, then they should probably be merged into one
\end{enumerate}
\item Can also include context from the other speaker (e.g., is one half of the utterance responding directly to the other speaker? It might make sense to split)
\end{enumerate}

Reminder: false starts with short gaps should be kept as one utterance (use “--”)

For long strings of repeated words or filler words (“yeah yeah yeah yeah ok”), follow the above rules, i.e., merge into one utterance unless intonation changes, or $\ge$ 2s pause

For ambiguous cases in whether to split into two utterances or keep as one, choose to split.

\subsection{Utterance timing }
Make sure the caption covers the entire span of the utterance (including any drawn-out endings of words). Note that the waveform won’t always provide enough information to determine when an utterance should start or end.

\subsection{Non-spoken noises}
Annotate:
\begin{itemize}
    \item Laughter, by itself – don’t annotate laughing while speaking
    \item Any noises that are unintentional but referred to (e.g., someone sneezes, the other person says “bless you”)
    \item Do not give utterance spans to other unintentional, non-spoken noises (e.g. coughs, hitting the microphone)
\end{itemize}
How to annotate:
\begin{itemize}
    \item  Put in \texttt{(())}. E.g., \texttt{((laughs))}
    \item Give it its own utterance span; separate from any spoken words.
\end{itemize}

\subsection{Unintelligible or partially intelligible segments}

\begin{itemize}
    \item If you have a decent guess on the interpretation, but uncertain, put in \texttt{( )}. E.g., \texttt{(on)}
    \begin{itemize}
        \item If there are two or more plausible interpretations, use / to separate. E.g., \texttt{(on/in)}
    \end{itemize}
    \item If you have no guess, use \texttt{()} with nothing inside (and no spaces). This can be used with an entire utterance, or just a part of an utterance if other parts are intelligible.
    \item For partially intelligible words, e.g., ``g–'' in a false start where it’s clear they meant ``go'', treat this as the first case, e.g., \texttt{(go)} and also mark as a false start, i.e., something like:
    \begin{itemize}
        \item \texttt{Let's (go) -\hspace{0pt}- let's go}
    \end{itemize}
\end{itemize}

\subsection{Intonation}
\begin{itemize}
    \item In general, we will not be annotating intonational features as part of the utterance annotation. E.g., rising intonation.
    \item We will be annotating timing information, e.g., pauses.
\end{itemize}

\subsection{Pauses and interruptions}
\begin{itemize}
    \item For false starts / self-correction/repair, use \texttt{-\hspace{0pt}-} with spaces surrounding.
    \item For pauses that are clearly pauses but <2 s:
    \begin{itemize}
        \item For short pauses (usually automatically annotated with \texttt{,} in Whisper), use \texttt{,}
        \item For longer pauses, use \texttt{...} with spaces surrounding.
    \end{itemize}
    \item For utterances that trail off or are interrupted:
    \begin{itemize}
        \item Always use \texttt{-\hspace{0pt}-} at the end.
        \item Don’t use ellipses at the end of any utterances.
    \end{itemize}
\end{itemize}

\subsection{Lengthened words}
Don’t indicate lengthening with punctuation.

\subsection{Punctuation}

In general, don’t bother changing Whisper punctuation too much unless it’s clearly incorrect about pauses.
However, don’t use \texttt{?} to mark uncertainty. Only use \texttt{?} for utterances that are syntactically questions (e.g., wh-questions, do-questions.)

\section{Data Post-Processing Guidelines}
\label{app:av_postprocessing}

To aid time-alignment, we configured the key-binding ``L'' to visually display a ``MARK'' message on-screen. This message event is also logged to the demo file.

The audio and video recordings were further post-processed.

\begin{enumerate}
    \item Using the on-screen marks, the recordings from the different players' perspectives were time-aligned. In the case that a mark was absent (e.g., if the player forgot to press ``L''), an in-game event (e.g., a button press or portal placement) was used to synchronize the timings. Recordings and demo files were time-aligned following the same procedure.
    \item Each of the audios were separately peak-normalized to 0.001dB so that the volume was consistent across players.
    \item Personally identifiable information was redacted in the transcript, and the corresponding section in the audio is replaced with white noise.
\end{enumerate}




\section{Dialogue acts}
\label{app:dialogue_acts}

The full set of dialogue acts used for both human and GPT-4o annotations, their definitions, and some examples are provided below.

\subsection{Communicative Status}

At a high level, communicative status categorizes whether a speaker's idea was successfully communicated. It distinguishes between fully conveyed thoughts, abandoned utterances, self-corrections, and unintelligible speech.

\paragraph{Uninterpretable}
 The utterance is severely fragmented, random, or semantically incoherent, making it impossible to interpret. If marked Uninterpretable, then Abandoned, Self-Correction, and Success cannot be marked.
\begin{itemize}
    \item \textbf{Examples:} \begin{itemize}
        \item  ``()'' (Empty utterance with no linguistic meaning.)     \end{itemize}

    \item \textbf{Non-Examples:}
    \begin{itemize}
        \item  ``I think—uh, never mind.'' (\textbf{Abandoned.} The intent is unclear but not entirely incomprehensible.)
        \item ``Oh, the door—wait, does it open from here?'' (\textbf{Self-Correction} - The speaker rephrases.)
        \item ``((laughter))'' (\textbf{Success} - The speaker successfully laughed.)
    \end{itemize}
\end{itemize}

\paragraph{Abandoned}
The utterance ends with ``--'', indicating an abandoned thought that is not completed later. An utterance is not abandoned if the speaker continues their thought after ``--'', even if short or fragmented.
\begin{itemize}
    \item \textbf{Examples:} \begin{itemize}
        \item ``Hey, I can take a look at—'' (No additional content follows.)
        \item ``So if you go through— you go— if I go—'' (\textbf{Abandoned \& Self-Correction.} Multiple phrase changes + trailing ``--''.)
    \end{itemize}
    \item \textbf{Non-Examples:}
    \begin{itemize}
        \item ``We just— you hit L.'' (\textbf{Correction.} The meaning switches from ``we just'' to ``You hit L''.)
        \item ``I don't see the— yeah.'' (\textbf{Correction.} The speaker changes from saying ``I don't see the'' to acknowledging the other speaker by saying ``yeah''.)
    \end{itemize}
\end{itemize}

\paragraph{Self-Correction}
The speaker modifies or switches meaning mid-sentence after ``--''. Both hesitation and rephrasing must be present to qualify as self-correction. Mere repetition or hesitation is \textbf{NOT} a self-correction.
\begin{itemize}
    \item \textbf{Examples:} \begin{itemize}
        \item  ``Take that left— actually, maybe right turn.'' (The meaning changes.)
        \item ``Wait, so where is— what is the objective here?'' (Reformulates inquiry.)
        \item  ``Oh no, but the— the— the target door did— the goal door did.'' (Last Hesitation Leads to a New Thought.)
    \end{itemize}
    \item \textbf{Non-Examples:}
    \begin{itemize}
        \item ``Yeah, this might all be uh— this might all be new.'' (\textbf{Success.} Hesitation, not correction.)
        \item ``I think I have to be outside to— to activate that door.'' (\textbf{Success.} No change in meaning, just a speech pause.)
    \end{itemize}
\end{itemize}

\paragraph{Success}
The utterance is clear, complete, and not marked as Uninterpretable, Abandoned or Self-Correction.
\begin{itemize}
    \item \textbf{Examples:} \begin{itemize}
        \item ``I think I have to be outside to— to activate that door.'' (Even though there's hesitation, the meaning is successfully conveyed.)
        \item ``That button turns off the turret.''
        \item ``I'll go first, then you follow.''
    \end{itemize}
\end{itemize}

\paragraph{Abandoned \& Self-Correction Together}
An utterance can be both \textbf{Abandoned} and \textbf{Self-Correction} if:
\begin{itemize}
    \item It ends with ``--'' (Abandoned).
    \item It contains multiple switches or rephrases mid-sentence (Self-Correction).
    \item \textbf{Example:} ``So if you go through— you go— if I go—'' (Has trailing ``--'' + multiple phrase changes. Both Abandoned and Self-Correction.)
\end{itemize}

\subsection{Utterance Type}

At a high level, utterance type categorizes spoken dialogue based on its function and intent in conversation. It systematically differentiates between statements, questions, commands, exclamations, and other speech acts based on their role in communication.

\paragraph{Proposition}
 The utterance makes a claim about the world that can be evaluated as true or false. To test, one can ask whether the utterance be followed by ``That's not true''.
\begin{itemize}
    \item \textbf{Examples:} \begin{itemize}
        \item ``There is a turret over there.''
        \item ``I think the exit is on the left.''
        \item ``The laser will activate the door.''
    \end{itemize}
    \item \textbf{Non-Examples:}
    \begin{itemize}
        \item ``Press the trigger.'' (\textbf{Imperative.} Command, not a statement.)
        \item ``Where do I place the portal?'' (\textbf{Query.} Asking, not claiming.)
    \end{itemize}
\end{itemize}

\paragraph{Repetition/Repair}
The speaker repeats or modifies a previous statement. If you believe this is a repetition/repair, also annotate the index of the utterance being repeated/modified.
\begin{itemize}
    \item \textbf{Examples:} \begin{itemize}
        \item Original: ``1. Blue: There is a turret over there.''
        \item Repetition/Repair: ``2. Orange: There is a blue turret over there.''
        \item Annotation: Repetition/Repair, 1.
    \end{itemize}
    \item \textbf{Non-Examples:}
    \begin{itemize}
        \item ``I saw a turret. Also, there's a button here.'' (\textbf{Proposition.} Adds new information rather than repeating.)
    \end{itemize}
\end{itemize}

\paragraph{Imperative}
The speaker is requesting or instructing an action, often beginning with a verb.
\begin{itemize}
    \item \textbf{Examples:} \begin{itemize}
        \item ``Press the trigger.''
        \item ``3, 2, 1, go.''
        \item ``You might want to put your blue portal a little higher.''
    \end{itemize}
    \item \textbf{Non-Examples:}
    \begin{itemize}
        \item ``Do you see the exit?'' (\textbf{Query.} Asking rather than instructing.)
        \item ``I think you should place it higher.'' (\textbf{Hedging} - Not a direct command.)
    \end{itemize}
\end{itemize}

\paragraph{Query}
The utterance seeks information, typically beginning with a question word (e.g., who, what, how, where, why).
\begin{itemize}
    \item \textbf{Examples:}
    \begin{itemize}
        \item ``Where do I place the portal?''
        \item ``Do you see the exit?''
        \item ``How does this switch work?''
    \end{itemize}
    \item \textbf{Non-Examples:}
    \begin{itemize}
        \item ``You placed the portal correctly, right?'' (\textbf{Tag Question.} A statement with a questioning phrase.)
        \item ``This thing takes in a laser, right?'' (\textbf{Tag Question.} Seeking confirmation rather than open-ended information.)
    \end{itemize}
\end{itemize}

\paragraph{Tag Question}
A statement modified by a phrase at the end, turning it into a question.
\begin{itemize}
    \item \textbf{Examples:}
    \begin{itemize}
        \item ``You placed the portal correctly, right?''
        \item ``We need to activate the switch, don't we?''
        \item ``This button controls the door, doesn't it?''
    \end{itemize}
    \item \textbf{Non-Examples:}
    \begin{itemize}
        \item ``How does this button work?'' (\textbf{Query.} A direct question, not a tag question.)
        \item ``I think this works.'' (\textbf{Proposition.} Statement without a questioning tag.)
    \end{itemize}
\end{itemize}

\paragraph{Exclamation or Performative}
The utterance either expresses strong emotion (Exclamation) or performs an action through speech (Performative).
\begin{itemize}
    \item \textbf{Examples:}
    \begin{itemize}
        \item ``Wow!''
        \item ``Oh no!''
        \item ``That was awesome!''
        \item ``I apologize for that.''
        \item ``I promise I'll go first.''
    \end{itemize}
    \item \textbf{Non-Examples:}
    \begin{itemize}
        \item ``That was really good.'' (\textbf{Proposition.} Evaluative but not an emotional outburst.)
        \item ``Yeah.'' (\textbf{Listener Response.} Simple acknowledgment.)
    \end{itemize}
\end{itemize}

\paragraph{Non-Sentential Utterances}
The utterance is not a full sentence but still conveys meaning.
\begin{itemize}
    \item \textbf{Examples:}
    \begin{itemize}
        \item ``One, three, four, and five.''
        \item ``Oh, yeah.''
        \item ``Nice, one more.''
        \item ``Uh-huh,'' ``Mhm,'' ``Right,'' ``Okay.'' (Backchanneling)
        \item ``I guess,'' ``Maybe,'' ``Hmm.'' (Minimal Agreement)
        \item ``Uh,'' ``Um,'' ``Erm.'' (Non-interruptive fillers)
    \end{itemize}
    \item \textbf{Non-Examples:}
    \begin{itemize}
        \item ``The exit is there.'' (\textbf{Proposition.} Complete statement.)
    \end{itemize}
\end{itemize}

\paragraph{Reported Speech}
The speaker reports speech, either directly or indirectly.
\begin{itemize}
    \item \textbf{Examples:}
    \begin{itemize}
        \item ``They said this button would open the door.''
        \item ``He told me to wait here.''
        \item ``She was like, `Let's go left.'''
    \end{itemize}
    \item \textbf{Non-Examples:}
    \begin{itemize}
        \item ``I think the button opens the door.'' (\textbf{Proposition.} Statement of belief, not reported speech.)
    \end{itemize}
\end{itemize}

\paragraph{Not Enough Information}
The utterance is incomplete or ambiguous, making it unclear what category applies.
\begin{itemize}
    \item \textbf{Examples:}
    \begin{itemize}
        \item ``I think—'' (Cut-off statement.)
        \item ``I can—'' (Incomplete thought.)
    \end{itemize}
    \item \textbf{Non-Examples:}
    \begin{itemize}
        \item ``I think the exit is on the left.'' (\textbf{Proposition.} Expressing belief.)
    \end{itemize}
\end{itemize}

\subsection{Information Level}

At a high level, information level captures the type of information conveyed in an utterance. It determines whether the speaker is describing the environment, explaining game mechanics, planning a task, managing conversation flow, expressing emotions, or discussing unrelated topics.

\paragraph{World State}
Describes the state of the game environment, including past, present, or future conditions. Focuses on confirming or describing static or dynamic world conditions. Unlike World Rules, World State is mutable—things in the world can change.
\begin{itemize}
    \item \textbf{Examples:}
    \begin{itemize}
        \item ``The door is open now.''
        \item ``Like—like this?'' (Confirming the state of the environment.)
        \item ``There's a laser pointing at the receiver.''
    \end{itemize}
    \item \textbf{Non-Examples:}
    \begin{itemize}
        \item ``This button opens the door.'' (\textbf{World Rules.} Describes a fixed game mechanic.)
        \item ``Stand on the button.'' (\textbf{Task-Related.} Command, not just describing.)
    \end{itemize}
\end{itemize}

\paragraph{World Rules}
Describes fixed mechanics of the game or invariant environmental rules, such as puzzle mechanics, door functions, laser interactions, or any consistent game behavior.
\begin{itemize}
    \item \textbf{Examples:}
    \begin{itemize}
        \item ``This thing takes in a laser, right?''
        \item ``Portals disappear when you walk through the fizzler.''
        \item ``The cube always respawns when it falls into the water.''
    \end{itemize}
    \item \textbf{Non-Examples:}
    \begin{itemize}
        \item ``There's a cube over there.'' (\textbf{World State.} Describing an object in the world, not a rule.)
        \item ``Try shining the laser here.'' (\textbf{Task-Related.} Suggesting an action.)
    \end{itemize}
\end{itemize}

\paragraph{Task-Related}
Involves actions, hypotheses, or strategy planning needed to solve the puzzle. Includes commands, task execution statements, and strategic reasoning.
\begin{itemize}
    \item \textbf{Examples:}
    \begin{itemize}
        \item ``Shine it on this, and I just have to time the jumps, right?''
        \item ``You go through first, and I'll follow.''
        \item ``I need to place my portal higher.''
    \end{itemize}
    \item \textbf{Non-Examples:}
    \begin{itemize}
        \item ``The door is open now.'' (\textbf{World State.} Describing, not planning.)
        \item ``Wait, does this button open the door?'' (\textbf{World Rules.} Asking about mechanics.)
        \item ``Sounds good.'' (\textbf{Communication Management.} Manages common ground by confirming; doesn’t add new task information.)
    \end{itemize}
\end{itemize}

\paragraph{Communication Management}
Ensures smooth conversation flow, understanding, or turn-taking. Includes acknowledgments, clarifications, and managing speech turns.
\begin{itemize}
    \item \textbf{Examples:}
    \begin{itemize}
        \item ``Yeah.'' (Simple acknowledgment.)
        \item ``Okay.'' (Confirming receipt of information.)
        \item ``Wait a minute.'' (Managing turn-taking.)
        \item ``I'm not sure.'' (Communicating uncertainty to a partner.)
    \end{itemize}
    \item \textbf{Non-Examples:}
    \begin{itemize}
        \item ``I think this works?'' (\textbf{Task-Related.} Expressing an action strategy.)
        \item ``That was amazing!'' (\textbf{Affective Evaluation.} Emotional expression.)
    \end{itemize}
\end{itemize}

\paragraph{Affective Evaluation}
Expresses emotion, appreciation, frustration, apologies, or personal evaluations. Captures positive and negative affect related to the game or interaction.
\begin{itemize}
    \item \textbf{Examples:}
    \begin{itemize}
        \item ``Great job!''
        \item ``Oops, my bad!''
        \item ``That was amazing!''
    \end{itemize}
    \item \textbf{Non-Examples:}
    \begin{itemize}
        \item ``Yeah, yeah, exactly.'' (\textbf{Communication Management.} Simple agreement without emotional tone.)
    \end{itemize}
\end{itemize}

\paragraph{Non-Task Related}
Talks about topics unrelated to solving the puzzle, such as off-task opinions, real-life events, and meta-discussion.
\begin{itemize}
    \item \textbf{Examples:}
    \begin{itemize}
        \item ``I wonder if people get through all four courses.''
        \item ``I'm hungry.''
        \item ``This reminds me of Portal 1.''
        \item ``We need to press L at the start of each level.'' (Referring to experiment setup.)
    \end{itemize}
    \item \textbf{Non-Examples:}
    \begin{itemize}
        \item ``We need to press the button to activate the door.'' (\textbf{Task-Related.} In-game action.)
    \end{itemize}
\end{itemize}

\paragraph{Not Enough Information}
The utterance lacks enough context to determine its category.
\begin{itemize}
    \item \textbf{Examples:}
    \begin{itemize}
        \item ``(What) --'' (Incomplete thought.)
        \item ``It's—'' (Cut-off statement.)
        \item ``Maybe we sh—'' (Unfinished suggestion.)
    \end{itemize}
    \item \textbf{Non-Examples:}
    \begin{itemize}
        \item ``Maybe we should press L?'' (\textbf{Task-Related.} Clear suggestion despite hedging.)
    \end{itemize}
\end{itemize}

\subsection{Uncertainty}

At a high level, uncertainty captures the degree of confidence the speaker expresses in their utterance. This can range from hedging (expressing uncertainty), certainty, or neutral statements that lack an explicit stance.

\paragraph{Hedging}
The speaker signals uncertainty, hesitation, or a lack of confidence.
\begin{itemize}
    \item Indicators include:
    \begin{itemize}
        \item Approximators: ``kind of,'' ``about,'' ``sometimes.''
        \item Hedges: ``probably,'' ``I think,'' ``maybe,'' ``it seems like.''
        \item Self-corrections with hesitation: ``Uh, where—wait, do you mean here?''
    \end{itemize}
    \item \textbf{Examples:}
    \begin{itemize}
        \item ``Uh, maybe we should press L?''
        \item ``I think the exit is over there, but I'm not sure.''
        \item ``It seems like this button controls the door?''
        \item ``I guess we need both cubes?''
    \end{itemize}
    \item \textbf{Non-Examples:}
    \begin{itemize}
        \item ``Where do we go?'' (\textbf{None.} Pure question, not a hedge.)
        \item ``Uh…'' (\textbf{None.} Filler, no hedging.)
    \end{itemize}
\end{itemize}

\paragraph{Certainty}
The speaker explicitly conveys high confidence in their statement.
\begin{itemize}
    \item Indicators include:
    \begin{itemize}
        \item Strong affirmations: ``I'm sure,'' ``Definitely,'' ``No doubt.''
        \item Absolute statements: ``It has to be this way,'' ``This is the only solution.''
    \end{itemize}
    \item \textbf{Examples:}
    \begin{itemize}
        \item ``I'm 100\% sure this button opens the door.''
        \item ``This is the only way to solve the puzzle.''
        \item ``That turret is definitely active.''
    \end{itemize}
    \item \textbf{Non-Examples:}
    \begin{itemize}
        \item ``Yeah.'' (\textbf{None.} Simple confirmation, no explicit certainty.)
        \item ``I think that works?'' (\textbf{Hedging.} Expresses uncertainty, not confidence.)
    \end{itemize}
\end{itemize}

\paragraph{Not Enough Information}
The utterance lacks enough context to determine certainty or uncertainty.
\begin{itemize}
    \item \textbf{Examples:}
    \begin{itemize}
        \item ``(What) --'' (Incomplete thought.)
        \item ``I think it's—'' (Cut-off statement.)
        \item ``Maybe we sh—'' (Hedge begins, but never finishes.)
    \end{itemize}
    \item \textbf{Non-Examples:}
    \begin{itemize}
        \item ``Uh, maybe we should press L?'' (\textbf{Hedging.} The uncertainty is clear.)
    \end{itemize}
\end{itemize}

\paragraph{None}
The utterance does not indicate certainty or uncertainty.
\begin{itemize}
    \item \textbf{Examples:}
    \begin{itemize}
        \item ``There's a button here.'' (Neutral statement, no confidence signal.)
        \item ``Press the switch.'' (Command, no uncertainty implied.)
        \item ``Do you see the exit?'' (Question without hedging.)
    \end{itemize}
    \item \textbf{Non-Examples:}
    \begin{itemize}
        \item ``I think the exit is over there.'' (\textbf{Hedging.} Expresses uncertainty.)
        \item ``I'm sure this is the way.'' (\textbf{Certainty.} Expresses confidence.)
    \end{itemize}
\end{itemize}

\subsection{Discursive Act}

At a high level, discursive act classification is about how utterances function within a collaborative task-based dialogue—specifically, how they contribute to decision-making, coordination, and knowledge exchange in a conversation.

\paragraph{Offer or Option}
The utterance includes a suggestion, proposal, or option.
\begin{itemize}
    \item \textbf{Examples:}
    \begin{itemize}
        \item ``We could try using the red portal instead.''
        \item ``Maybe we should check behind the door first.''
        \item ``You might want to stand on that button while I go through.''
    \end{itemize}
    \item \textbf{Non-Examples:}
    \begin{itemize}
        \item ``Stand on the button.'' (\textbf{Directive.} Command.)
        \item ``The button opens the door.'' (\textbf{Assertion.} Statement of fact.)
    \end{itemize}
\end{itemize}

\paragraph{Directive}
A request or command for action.
\begin{itemize}
    \item \textbf{Examples:}
    \begin{itemize}
        \item ``Press the switch now.''
        \item ``Step on the pressure plate while I go ahead.''
        \item ``Put your portal on the white surface to the left.''
    \end{itemize}
    \item \textbf{Non-Examples:}
    \begin{itemize}
        \item ``Maybe you should press the switch?'' (\textbf{Offer or Option.} Suggestion.)
    \end{itemize}
\end{itemize}

\paragraph{Request for Information}
A request for new details about the game.
\begin{itemize}
    \item \textbf{Examples:}
    \begin{itemize}
        \item ``Does stepping on this button open the door?''
        \item ``What does that switch do?''
        \item ``How many turrets are in this room?''
    \end{itemize}
    \item \textbf{Non-Examples:}
    \begin{itemize}
        \item ``Is that a blue turret or a red turret?'' (\textbf{Request for Clarification.} Clarifying known details.)
    \end{itemize}
\end{itemize}

\paragraph{Request for Clarification}
Seeks to resolve ambiguity about a previous utterance.
\begin{itemize}
    \item \textbf{Examples:}
    \begin{itemize}
        \item A: ``We need to activate the switch.'' \\ B: ``Which switch are you talking about?''
        \item A: ``You should go first.'' \\ B: ``Do you mean through the door or across the bridge?''
    \end{itemize}
    \item \textbf{Non-Examples:}
    \begin{itemize}
        \item ``Where is the switch?'' (\textbf{Request for Information.} Seeking new knowledge.)
    \end{itemize}
\end{itemize}

\paragraph{Assertion}
Describes the world state or game environment.
\begin{itemize}
    \item \textbf{Examples:}
    \begin{itemize}
        \item ``There's a turret behind the wall.''
        \item ``The door only opens if both of us stand on the pressure plates.''
        \item ``This room is identical to the last one.''
    \end{itemize}
    \item \textbf{Non-Examples:}
    \begin{itemize}
        \item ``There might be a turret behind the wall.'' (\textbf{Speculation.} Hypothetical guess.)
    \end{itemize}
\end{itemize}

\paragraph{Justification}
Elaborates on a previous utterance by providing reasoning.
\begin{itemize}
    \item \textbf{Examples:}
    \begin{itemize}
        \item ``We should use the portal here because it gives us a clear shot at the exit.''
        \item ``Let's avoid that path since there are too many turrets.''
        \item ``I placed my portal lower so that we have an easier jump.''
    \end{itemize}
    \item \textbf{Non-Examples:}
    \begin{itemize}
        \item ``We should use the portal here.'' (\textbf{Offer or Option.} Standalone suggestion.)
    \end{itemize}
\end{itemize}

\paragraph{Speculation}
A guess or hypothesis about the game state.
\begin{itemize}
    \item \textbf{Examples:}
    \begin{itemize}
        \item ``I think the exit is behind that door.''
        \item ``Maybe we need both cubes to solve the puzzle?''
        \item ``I guess that laser might deactivate the turrets.''
    \end{itemize}
    \item \textbf{Non-Examples:}
    \begin{itemize}
        \item ``The exit is behind that door.'' (\textbf{Assertion.} Statement of fact.)
    \end{itemize}
\end{itemize}

\paragraph{Commit}
Accepts a previous request or offer.
\begin{itemize}
    \item \textbf{Examples:}
    \begin{itemize}
        \item A: ``Can you stand on the button?'' \\ B: ``Sure, I'll do it now.''
        \item A: ``Should we try going through the left door?'' \\ B: ``Yeah, let's do that.''
    \end{itemize}
    \item \textbf{Non-Examples:}
    \begin{itemize}
        \item ``Okay.'' (\textbf{Acknowledgment.} Could be a general response.)
    \end{itemize}
\end{itemize}

\paragraph{Confirmation or Status Marker}
Signals task completion or status change.
\begin{itemize}
    \item \textbf{Examples:}
    \begin{itemize}
        \item ``I've placed my portal.''
        \item ``The door is open now.''
        \item ``I'm ready when you are.''
    \end{itemize}
    \item \textbf{Non-Examples:}
    \begin{itemize}
        \item ``Nice, one more to go.'' (\textbf{Acknowledgment.} Recognizes progress but not a status change.)
    \end{itemize}
\end{itemize}

\paragraph{Acknowledgment}
Confirms or acknowledges a previous utterance.
\begin{itemize}
    \item \textbf{Examples:}
    \begin{itemize}
        \item ``Got it.''
        \item ``Makes sense.''
        \item ``Okay, that helps.''
    \end{itemize}
    \item \textbf{Non-Examples:}
    \begin{itemize}
        \item ``Yeah, let's do that.'' (\textbf{Commit.} Agreement and acceptance of a proposal.)
    \end{itemize}
\end{itemize}

\paragraph{Rejection}
Disagrees with or rejects a previous utterance.
\begin{itemize}
    \item \textbf{Examples:}
    \begin{itemize}
        \item A: ``Let's go left.'' \\ B: ``No, I think right is safer.''
        \item A: ``You should press the button first.'' \\ B: ``I don't think that's necessary.''
    \end{itemize}
    \item \textbf{Non-Examples:}
    \begin{itemize}
        \item ``I'm not sure if that will work.'' (\textbf{Speculation.} Hesitation rather than direct rejection.)
    \end{itemize}
\end{itemize}

\paragraph{Expressive}
The utterance expresses an emotional reaction, interjection, or spontaneous emotional vocalization that is not part of task-related reasoning, but still contributes socially or emotionally to the interaction. This captures non-strategic emotional outbursts such as frustration, surprise, laughter, or celebration, that aren't evaluative (like ``great job'') or affectively reasoned.
\begin{itemize}
    \item \textbf{Examples:}
    \begin{itemize}
        \item ``Shoot!'' (Frustration after making a mistake.)
        \item ``Ugh!'' (Annoyance.)
        \item [laughs] (Laughter as a standalone reaction.)
        \item ``Woo!'' (Celebration.)
        \item ``Yikes!'' (Startled reaction.)
    \end{itemize}
    \item \textbf{Non-Examples:}
    \begin{itemize}
        \item ``No, I think we should go right.'' (\textbf{Rejection.} Disagrees with a previous proposal.)
        \item ``I'm not sure if that'll work...'' (\textbf{Speculation.} Offers a task-relevant hypothesis.)
        \item ``That button controls the turret.'' (\textbf{Assertion.} Statement about the game state.)
    \end{itemize}
\end{itemize}

\paragraph{Not Enough Information}
The utterance lacks enough context for classification.
\begin{itemize}
    \item \textbf{Examples:}
    \begin{itemize}
        \item ``Uh—maybe we should—''
        \item ``I think it's—never mind.''
        \item ``Well, we could… hmm…''
    \end{itemize}
    \item \textbf{Non-Examples:}
    \begin{itemize}
        \item ``Let's go left.'' (\textbf{Directive.} Clear instruction.)
    \end{itemize}
\end{itemize}

\end{document}